# Clustering Text Using Attention

Lovedeep Singh
*Computer Science and Engineering*
*Punjab Engineering College*
Chandigarh, India
masterlovedeep.singh97@gmail.com

*Abstract*—Clustering Text has been an important problem in the domain of Natural Language Processing. While there are techniques to cluster text based on using conventional clustering techniques on top of contextual or non-contextual vector space representations, it still remains a prevalent area of research possible to various improvements in performance and implementation of these techniques. This paper discusses a novel technique to cluster text using attention mechanisms. Attention Mechanisms have proven to be highly effective in various NLP tasks in recent times. This paper extends the idea of attention mechanism in clustering space and sheds some light on a whole new area of research

## I. INTRODUCTION

There are various situations where the need is to group similar texts into same buckets. We do not have enough previous experience or knowledge to run a classification algorithm on top of the available data. Clustering is the fundamental and intuitive solution to such problems. Text clustering is more challenging than normal feature dataset clustering since it requires prior feature extraction or some sort of mathematical treatment before passing it to a clustering algorithm. The pre algorithm treatment done on text either to get some vector representation or a feature set for a given piece of text plays a significant part in the ultimate performance of the clustering technique. This complete pipeline of pre-algorithm treatment opens up possibilities of various permutations and combinations apart from the final clustering algorithm to improve overall performance. Attention mechanism has already taken over NLP research like a big wave resulting in more advances and possibilities in this area. In this paper, we discuss the use of attention mechanism in this pipeline to improve the overall performance of clustering.

## II. RELATED WORK

Aggarwal and Zhai [1] have comprehensively covered the concepts and practices in text clustering, both in terms of pre-algorithm treatment and different clustering techniques. Authors have shed light on most of the important ways to cluster text. Jianping et al. [2] have also discussed more advanced techniques which leverage heterogeneous information networks [3] and graphical analysis to perform text clustering. These techniques based on network structure lie bit in a different area than the vector representation techniques used in pre-algorithm part. Grzegorczyk [4], Babic et. al. [5] provide a holistic and in depth view in the various text representation techniques in vector space and their modelling.

The intuition for attention mechanism and the basic idea behind the same has been existing in the field of deep learning from quite a bit of time. It came to the field of Computer Vision first before diffusing in other areas. It was first introduced by Bahdanau et al. [6] in the field of NLP. Even though the concept of attention mechanism has been long prevailing, the major attention of the researchers towards attention mechanism arose with the significant results and contributions by Vaswani et. al. [7] which introduced the Transformers architecture to the NLP field. Although transformers were initially proposed for Neural Machine Translation, the pipeline discussed in the paper is quite deep and novel in its own sense. The individual encoder and decoder parts have given BERT [8] and GPT [9], one of the most advanced language models to the world respectively. Galassi et. al. [10] have done a great work and given a comprehensive, structured, in-depth, and sound analysis of attention in natural language processing.

## III. ATTENTION MECHANISMS

In simple terms, attention mechanism can be thought of an additional layer somewhere in a network architecture which gives the deep learning model extra controlling parameters to refine its learning by paying attention to different parts of the input as per the requirement.

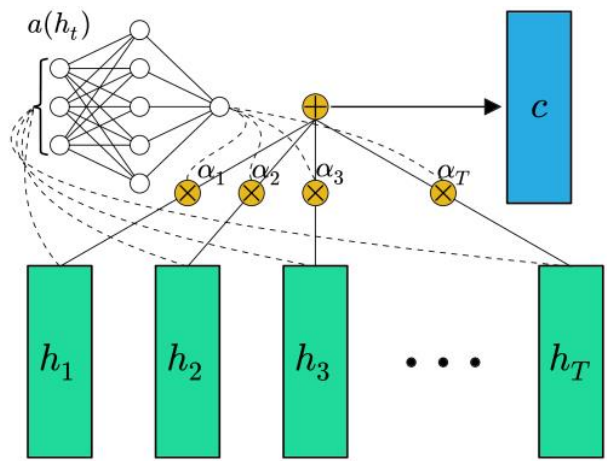

Fig. 1. Simple Feed Forward Attention Mechanism

Attention mechanism can be a simple feed forward neural network in the deep learning model which provides additional parameters (the weights the model learns), we can call these as attention weights. Each weight reflects the amount of attention the models is paying to a specific part of the input to reach the output. Fig 1. depicts attention mechanism used in a simple feed forward neural network. This has been mentioned by Colin et. al. [11] where they support that feed forward neural network with attention can solve some long term memory problems.

Attention mechanisms when combined with Vanilla RNN based encoder decoder models gives better performance in Neural Machine Translation (NMT) tasks. Vaswani et. al. [7] went to the next level by removing RNNs completely from NMT architectures and using only attention for NMT. In the paper, Attention is all you need [7], the authors incorporated some effect of RNNs by concatenating positional embeddings along with the word embeddings. The attention mechanism discussed in form of Key, Query and Vector also strengthens the belief of giving extra handles to model to leverage upon can increase the model performance.

Hierarchical Attention Networks, proposed by Zang et. al. [12] proposes two levels of attention for document classification tasks. It uses attention mechanism in an hierarchical fashion and leverages attention at both word and sentence level before a final softmax layer for classifying the document. This provides the model opportunity to pay attention to important words in a sentence as well as important sentences in a document. This opportunity comes with a cost, model has to learn additional parameters now so as to leverage such a hierarchical structure, but the performance improvement is much more than the cost of additional parameters.

## IV. CLUSTERING TECHNIQUES

Most of the clustering techniques leverage distance metric to partition data into clusters after representing data in a n-dimensional hypercube. During clustering text, we can control the variable "n", the dimensions of our input vector representation of text. It is quite possible that varying the "n"., i.e., the dimensions of the input vector, we may get different results for different clustering algorithms. Many clustering algorithms also require number of clusters as an input. If our experience or knowledge about data is limited, estimating number of clusters can be challenging. Though there are techniques available to estimate number of clusters, usually Elbow method or Silhouette score analysis. Sometimes, researchers also use square-root of number of data points as an estimate for number of clusters.

## V. PROPOSED METHOD

We use Hierarchical Attention in the pre-algorithm

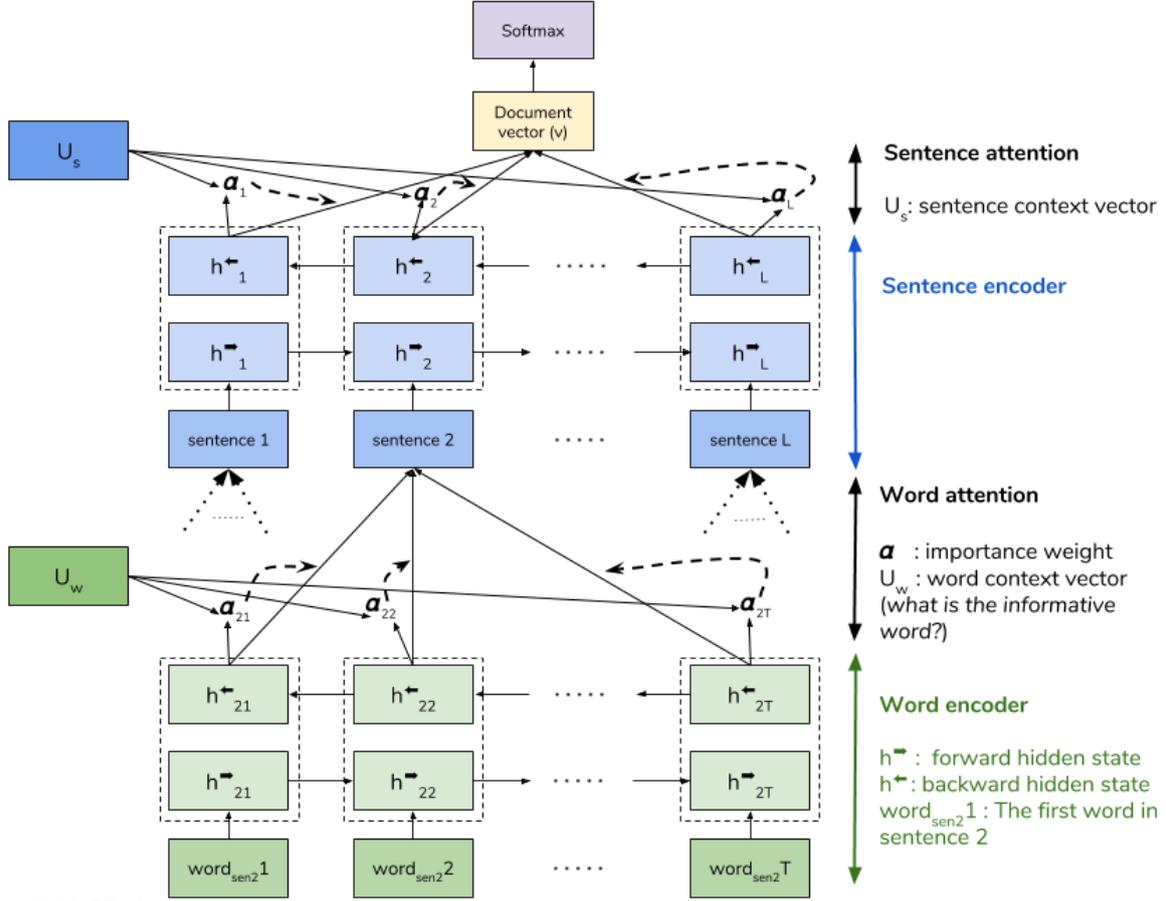

Fig. 2. Hierarchical Attention Network for document classification

treatment before application of any clustering technique. The intuition is taken from the work done on document classification using Hierarchical Attention Networks by Zang et. al. [12].

Fig 2. clearly depicts the network architecture used in Hierarchical Attention Networks for document classification. Each document is made of a number of sentences and each sentence made of a number of words. We leverage the same architecture to get the Document Vector and use it as input vector for the clustering techniques.

Words are the smallest atomic unit in our architecture. Each word is represented using a fixed dimension embedding. These words in the same order as they appear are passed to a Bi-LSTM layer to incorporate the effect of positioning of words with respect to its surroundings. Following the bidirectional LSTM is an attention layer for word level attention. After this layer we get sentence context vector, a representation for sentence with attention applied at word level. Many such sentences combine to form a document. We repeat the same Bi-LSTM followed by attention layer to get the Document Vector, a vector representation for document which can be used as an input for clustering algorithm.

The weights in the network architecture in original classification task are learned while training and these weights help to generate meaningful document vectors before a final softmax layer while testing. Since clustering is an unsupervised algorithm, it does not have any training before testing. How do we get weights in different layers to get document vector before passing it to clustering ? We use classification problem to learn these weights before jumping into clustering.

The concept used to learn the weights goes as follows. During preliminary data analysis, we separate out some fraction of data from the complete data. We perform manual annotation on this fractional data and segregate it into different classes.Then, we train a classification model using Hierarchical Attention Network on this fractional data. This training helps us to learn weights and get insight into the structure and partitioning present in the data. Finally we save these learned parameters by our model. We use these learned parameters in order to get document vectors of remaining fraction of data before passing it to a clustering algorithm.

The intuitive idea behind this reflects the real world scenario. In real world scenarios, whenever we are faced with some challenges which require bucketization of data, we usually look at the data and analyse it before jumping on to the clustering techniques. In this preliminary data analyses, the data scientist tries to manually annotate some fraction of data to capture various parameters, one important parameter being the number of clusters. Since the data scientist is already performing such an analyses, a little more effort in this direction to annotate fraction of data into separate classes could help us learn attention weights and other model parameters of Hierarchical Attention Networks which could be leveraged to improve performance of clustering algorithms.

## VI. Experiments and Results

We perform a number of experiments in which we change either word embeddings, initial fraction of data to learn weights, or the clustering technique. We experiment with both pre-trained word embeddings and word embeddings[1] trained using the actual data.

We also perform number of experiments for plain clustering by varying clustering algorithms. In plain clustering, we use Doc2vec[2] to represent a document in vector space before passing it to clustering algorithm.

We have used the drugs dataset [13] in our experiments. This is a tabular dataset. We use "review" as input and "condition" as the output for training attention weights during pre-clustering classification. We filter out conditions with less than 3 entries and kept max 20 entries for a condition. We split the dataset into training set and clustering set evenly using stratification. We pass number of classes as number of clusters where need.

We define Avg evaluation metric Avg.Ev. to plot performance of various clustering algorithms.

$$Avg.Ev. = (homo+comp+var+ari+ami+silh)/6$$

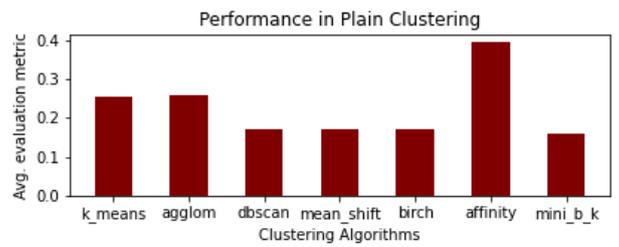

Fig. 3. Performance bar chat of Avg.Ev. in Plain Clustering

Table 1. describes all the variations we try in our experiments.

TABLE I. VARIATIONS

| Variation Code | Variation meaning |
|---|---|
| ASn | Attention Clustering with selftrained word embeddings and using n/10 fraction of data for pre-clustering classification training |
| APn | Attention Clustering with pre trained word embeddings and using n/10 fraction of data for pre-clustering classification training |
| AP | Attention Clustering with pre trained word embeddings and using some variable fraction of data for preclustering classification training |
| AS | Attention Clustering with selftrained word embeddings and using some variable fraction of data for preclustering classification training |

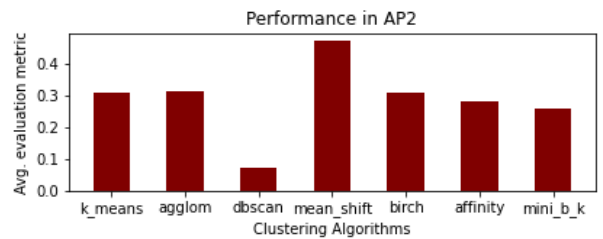

Fig. 4. Avg.Ev. bar chart for Variation Code AP2

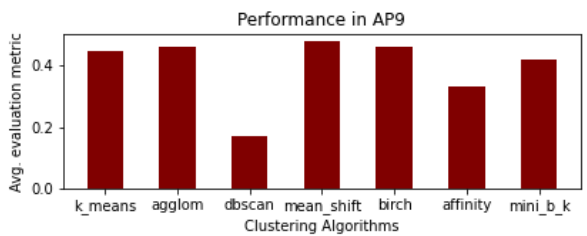

Fig. 5. Avg. Ev. Bar chart for Variation Code AP9

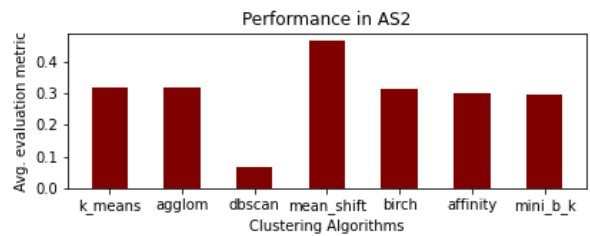

Fig. 6. Avg. Ev. Bar chart for Variation Code AS2

---

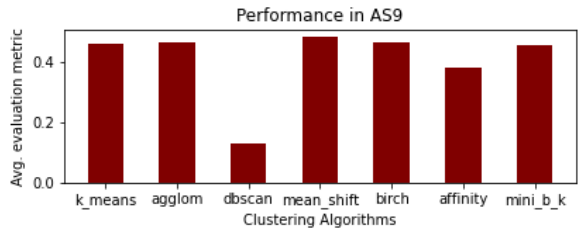

Fig. 7. Avg. Ev. Bar chart for Variation Code AS9

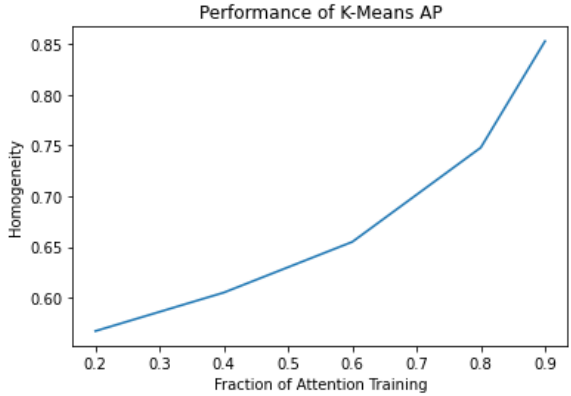

Fig. 8. Line Chart of Homogeneity score in K-Means with Variation AP

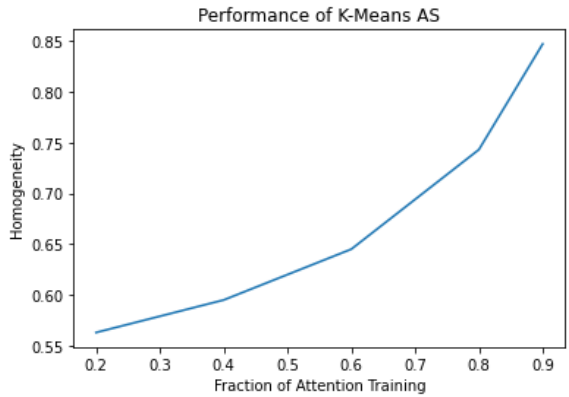

Fig. 9. Line Chart of Homogeneity score in K-Means with Variation AS

TABLE II. PLAIN CLUSTERING RESULTS

| Algorithm | Homo | Comp | V-me | ARI | AMI | Silh |
|---|---|---|---|---|---|---|
| k-means | .498 | .514 | .506 | .000 | .002 | .008 |
| agglom | .502 | .516 | .509 | .000 | .002 | .012 |
| dbscan | .000 | 1.000 | .000 | .000 | .000 | ---- |
| meanshift | .000 | 1.000 | .000 | .000 | .000 | ---- |
| birch_fn | .000 | 1.000 | .000 | .000 | .000 | ---- |
| affinity | .917 | .659 | .767 | .001 | .003 | .018 |
| minibkmea | .277 | .386 | .332 | .000 | .004 | -.06 |

TABLE III. AP2 COMPLETE RESULTS

| Algorithm | Homo | Comp | V-me | ARI | AMI | Silh |
|---|---|---|---|---|---|---|
| k-means | .567 | .580 | .574 | .015 | .057 | .041 |
| agglom | .580 | .587 | .583 | .017 | .062 | .039 |
| dbscan | .002 | .566 | .004 | .000 | .000 | -.15 |
| meanshift | .999 | .706 | .827 | .029 | .053 | .211 |
| birch_fn | .577 | .585 | .581 | .017 | .061 | .038 |
| affinity | .513 | .558 | .535 | .013 | .054 | .023 |
| minibkmea | .471 | .539 | .503 | .010 | .050 | -.04 |

TABLE IV. AP9 COMPLETE RESULTS

| Algorithm | Homo | Comp | V-me | ARI | AMI | Silh |
|---|---|---|---|---|---|---|
| k-means | .853 | .893 | .873 | .007 | .010 | .041 |
| agglom | .880 | .896 | .888 | .010 | .011 | .074 |
| dbscan | .000 | 1.000 | .000 | .000 | .000 | ---- |
| meanshift | .999 | .906 | .950 | .005 | .002 | .005 |
| birch_fn | .880 | .896 | .888 | .010 | .011 | .074 |
| affinity | .525 | .840 | .646 | .001 | .004 | .020 |
| minibkmea | .799 | .887 | .841 | .003 | .007 | -.02 |

TABLE V. AS2 COMPLETE RESULTS

| Algorithm | Homo | Comp | V-me | ARI | AMI | Silh |
|---|---|---|---|---|---|---|
| k-means | .563 | .578 | .570 | .012 | .051 | .143 |
| agglom | .562 | .577 | .569 | .012 | .052 | .145 |
| dbscan | .025 | .456 | .047 | .000 | .002 | -.14 |
| meanshift | .962 | .698 | .809 | .023 | .053 | .245 |
| birch_fn | .554 | .575 | .564 | .011 | .051 | .140 |
| affinity | .521 | .560 | .564 | .010 | .047 | .136 |
| minibkmea | .518 | .559 | .537 | .009 | .047 | .097 |

TABLE VI. AS9 COMPLETE RESULTS

| Algorithm | Homo | Comp | V-me | ARI | AMI | Silh |
|---|---|---|---|---|---|---|
| k-means | .847 | .891 | .869 | .002 | .003 | .135 |
| agglom | .853 | .892 | .872 | .002 | .003 | .157 |
| dbscan | .030 | .775 | .059 | .000 | .000 | -.09 |
| meanshift | .973 | .904 | .937 | .001 | .001 | .072 |
| birch_fn | .850 | .892 | .870 | .002 | .003 | .156 |
| affinity | .620 | .859 | .720 | .001 | .002 | .087 |
| minibkmea | .850 | .891 | .870 | 001 | .001 | .119 |

## VII. DISCUSSION

We can see some general trends in experiments. In general, attention clustering performed better than plain clustering. Within attention clustering, results of both variation AS and AP were not very different, i.e., with both pre trained word embeddings and self-trained word embeddings, we got similar results. This is because drugs dataset has a lot of common English words, even though it has some specific jargon in it, pre trained word embeddings were still able to capture that. Additionally, attention weights helped the clustering algorithm in capturing the differences in both AP and AS variations almost equally well. This leaves an intuitive suggestion that even random vector space representations without any contextual meaning attached may give some reasonable performance rather than completely flawed performance metrics because of attention training involved in the pre-clustering pipeline.

Also within the attention clustering variations, in both AS and AP, clustering algorithm performance metric improved with increase in fraction of data used for attention training. This is quite intuitive, as more data was used for attention training, the better the attention weights were able to capture the signal present in the data.

We have provided result tables, bar charts and line charts for some limited variations due to space constraints. Results of all other variations can be found in the official code of this research work. All the code is available at GitHub[3] for further research and experimentation.

## VIII. CONCLUSION

It is evident that clustering using attention mechanism indeed help in the overall performance of the clustering algorithm. The performance improves with increase in fraction of data used for attention training. We have used Hierarchical Attention Networks for our experiment, there could be other ways to incorporate attention mechanism in the pre-clustering pipeline. Self-attention and attention used in Transformers could be another possible way. This paper tries to shed light into the less explored possibilities in the clustering field.

## REFERENCES

[1] Aggarwal C.C., Zhai C. (2012) A Survey of Text Clustering Algorithms. In: Aggarwal C., Zhai C. (eds) Mining Text Data. Springer, Boston, MA. https://doi.org/10.1007/978-1-4614-3223-4_4

[2] Cao, Jianping & Wang, Senzhang & Wen, Danyan & Peng, Zhaohui & Yu, Philip. (2019). Mutual Clustering on Comparative Texts via Heterogeneous Information Networks.

[3] C. Shi, Y. Li, J. Zhang, Y. Sun and P. S. Yu, "A survey of heterogeneous information network analysis," in IEEE Transactions on Knowledge and Data Engineering, vol. 29, no. 1, pp. 17-37, 1 Jan. 2017, doi: 10.1109/TKDE.2016.2598561.

[4] Karol Grzegorczyk. (2019). Vector representations of text data in deep learning. arXiv:1901.01695

[5] K. Babić, S. Martinčić-Ipšić, and A. Meštrović, "Survey of Neural Text Representation Models," Information, vol. 11, no. 11, p. 511, Oct. 2020 [Online]. Available: http://dx.doi.org/10.3390/info11110511

[6] D. Bahdanau, K. Cho, and Y. Bengio, "Neural machine translation by jointly learning to align and translate," in Proc. ICLR, 2015, pp. 1–15.

[7] Ashish Vaswani, Noam Shazeer, Niki Parmar, Jakob Uszkoreit, Llion Jones, Aidan N. Gomez, Łukasz Kaiser, and Illia Polosukhin. 2017. Attention is all you need. In Proceedings of the 31st International Conference on Neural Information Processing Systems (NIPS'17). Curran Associates Inc., Red Hook, NY, USA, 6000–6010.

[8] Devlin, J., Chang, M., Lee, K., & Toutanova, K. (2019). BERT: Pretraining of Deep Bidirectional Transformers for Language Understanding. NAACL-HLT.

[9] Radford, A., & Narasimhan, K. (2018). Improving Language Understanding by Generative Pre-Training.

[10] A. Galassi, M. Lippi and P. Torroni, "Attention in Natural Language Processing," in IEEE Transactions on Neural Networks and Learning Systems, doi: 10.1109/TNNLS.2020.3019893.

[11] Raffel, Colin & Ellis, Daniel. (2015). Feed-Forward Networks with Attention Can Solve Some Long-Term Memory Problems.

[12] Yang, Zichao & Yang, Diyi & Dyer, Chris & He, Xiaodong & Smola, Alex & Hovy, Eduard. (2016). Hierarchical Attention Networks for Document Classification. 1480-1489. 10.18653/v1/N16-1174

[13] Felix Gräßer, Surya Kallumadi, Hagen Malberg, and Sebastian Zaunseder. 2018. Aspect-Based Sentiment Analysis of Drug Reviews Applying Cross-Domain and Cross-Data Learning. In Proceedings of the 2018 International Conference on Digital Health (DH '18). ACM, New York, NY, USA, 121-125. DOI: https://doi.org/10.1145/3194658.3194677

---

[3] https://github.com/singh-l/CTUA